\documentclass[a4paper]{article}
\usepackage[it]{idsiatechrep}   

\usepackage{times}
\usepackage{graphicx}
\usepackage{booktabs}
\usepackage{tabularx}

\title{Multi-Column Deep Neural Networks\\ for Offline Handwritten Chinese Character Classification}
\author{Dan Cire\c{s}an and J\"{u}rgen Schmidhuber}
\date{August 30, 2013}

\reportnumber{05-13}

\reportaddnote{This work was partially supported by the Supervised Deep / Recurrent Nets SNF grant, Project Code
140399.}

\begin{document}
\makecover         
\maketitle

\begin{abstract}
Our Multi-Column Deep Neural Networks achieve best known recognition rates on Chinese characters from the ICDAR 2011 and 2013 offline handwriting competitions, approaching human performance.
\end{abstract}

\section{Introduction}

Deep and wide max-pooling convolutional neural networks (MPCNN) on GPU \cite{Ciresan:2011a} embody the current state of the art in stationary pattern recognition. They outperformed other methods on  image classification \cite{Ciresan:2012b}, object detection\cite{Ciresan:2013miccai}, and image segmentation  \cite{Ciresan:2012f, farabet-pami-13}.  
Through output averaging, several independently trained deep NN (DNN) can form a Multi-Column DNN (MCDNN) with error rates 20-40\%  below those of single DNN \cite{Ciresan:2012b}.

In 2012, our MCDNN were the first to achieve first human-competitive performance on the famous MNIST handwritten digit recognition task, e.g., \cite{Ciresan:2012b}. 
Chinese handwriting, however, is much harder, as there are not only 10 classes (one for each digit), but 3755. 

Here we apply our MCDNN to data from the ICDAR 2013 competition \cite{icdar2013Chinese} on recognizing offline handwritten Chinese characters. 
We present results obtained after correcting a bug (Section~\ref{sec:bug}) in the image preprocessing routine. 

\section{Details}
We use several MCDNN architectures to classify handwritten Chinese characters from the dataset used at ICDAR 2011 \cite{icdar2011Chinese} and 2013 \cite{icdar2013Chinese} competitions.
All training was done prior to the competition deadline. An executable was submitted to the organizers, whose test set was released after the 2013 competition, allowing us to further verify our MCDNN.
\subsection{Data}
Details can be found in the competition reports \cite{icdar2011Chinese,icdar2013Chinese}. The data consists of plain images (offline, no temporal information) of isolated  Chinese characters (already segmented out from text). The test set was identical for both the 2011 and 2013 competitions. It contains 224419 characters written by 60 persons.

Dataset  HWDB 1.1 contains characters written by 240 persons for actual training and by 60 for validation:  897758 and 223991 characters, respectively.
Note that there are far more classes (3755) than samples per class (240+60).

\subsection{Preprocessing}
Although Chinese has tens of thousands of different classes,  HWDB 1.1 \cite{icdar2011Chinese} contains only the 3755 most frequent ones. They require more complicated graphics than the 26 classes of Latin letters. This requires bigger images. Our experience with handwritten digits and Latin letters \cite{Ciresan:2011c,Ciresan:2012a} tells us that a $20\times20$ pixel rectangular image can show enough details for good recognition. After visual inspection of several Chinese characters rescaled to various sizes we decided on using $40\times40$ pixel images. Scaling is done uniformly; the biggest dimension of each character determines the scaling factor. We also place scaled characters in the middle of $48\times48$ pixel images, to allow for various deformations during the training process. Before resizing, we maximize input image contrast to get values from 0 to 255.

\subsection{Preprocessing glitch at ICDAR}
\label{sec:bug}
Our training and testing framework \cite{Ciresan:2011a} is designed for already preprocessed data, that is, neither training nor testing involves preprocessing. Instead, dedicated Matlab programs are used to preprocess data whenever necessary. For Chinese characters, preprocessing is limited to rescaling the images to a fixed size, plus simple contrast maximization.

\begin{figure}
\includegraphics[width=\textwidth]{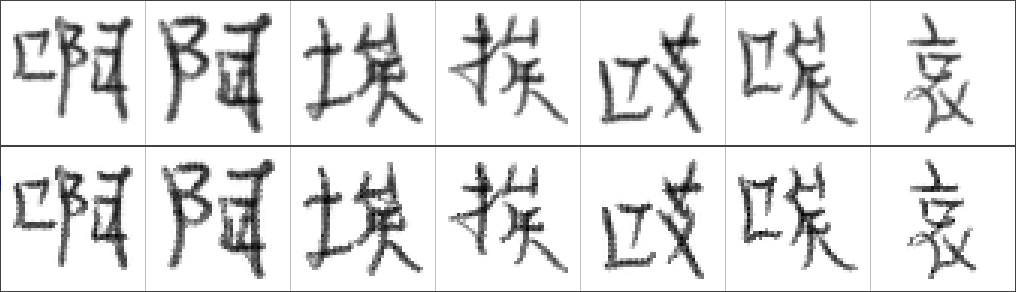}
   \caption{First seven characters of the competition test set preprocessed in Matlab (\emph{first row}) and OpenCV (\emph{second row}).}
\label{fig:preprocessingbug}
\end{figure}

ICDAR requires executables, hence we rewrote preprocessing routines in C++, using the OpenCV library instead of writing a new scaling function. As we learned the hard way, however, Matlab and OpenCV scaling routines do not produce exactly the same results (Fig.~\ref{fig:preprocessingbug}), despite using the same interpolation method. Characters in Fig.~\ref{fig:preprocessingbug} look alike, but the ones preprocessed in OpenCV are much grainier. Our executable also reversed the order of scaling and contrast maximization. As a consequence, our framework was trained and validated with one preprocessing routine, while the submitted executable used a different one. Since the feedback from the organizers matched our expectations, the problem was noticed only once the test data was released after the competition. 

When we applied identical preprocessing for both training and test set, the test error was 4.21\%, down from our original competition result of 5.58\%. All our DNN and MCDNN had their error rates reduced, by up to 2\%. 

Since we also submitted  an executable with the same flawed preprocessing to the 2011 competition using the same test data, we rechecked the 2011 result, and also got a 2.04\% lower error rate (5.78\% instead of 7.82\%).

Despite flawed preprocessing we won the 2011 competition. But we lost the 2013 competition by 0.35\%, coming in 2nd at 5.58\% vs. 5.23\%. With correct preprocessing, however, we get 1.01\% absolute error rate reduction (a massive 19.3\% in relative reduction) over the team which ranked first.

\subsection{Network architecture}
We train eight networks (Table~\ref{tab:net_architectures}) on HWDB1.1. All networks have 11 layers, counting input and output layers. The number of maps per layer varies from 100 to 450. We also try two different sizes for the first fully connected layer. The last layer always has 3755 neurons, i.e. one per class. The last four nets are trained on the HWDB 1.1 training set, i.e., characters written by 240 persons. The first four nets are trained on characters written by all 300 persons associated with training and validation datasets.

\begin{table}[h!]
\caption{Network architectures. 48x48 represents input layer size, xCy a convolutional layer with x maps and filters of $y\times{y}$ weights, MPy  a max-pooling layer with $y\times{y}$ pooling size, xFC a fully connected layer with x neurons. The code suffix uniquely identifies a trained network. The last but one column shows the errors of individual DNN. Speed in ms per character (on NVIDIA GTX 580) is displayed in the last column.
\label{tab:net_architectures}} 
\begin{center}
\footnotesize
\begin{tabular}{c|lrr}
\toprule
\#	&	Architecture			&	Error[\%]	&	Speed [ms/character]\\
\midrule
0		&	48x48-150C3-MP2-250C2-MP2-350C2-MP2-450C2-MP2-1000N-3755N-1365334845	&	5.528	&	3.03\\	
1		&	48x48-150C3-MP2-250C2-MP2-350C2-MP2-450C2-MP2-1000N-3755N-1365775809	&	5.931	&	3.03\\
2		&	48x48-300C3-MP2-300C2-MP2-300C2-MP2-300C2-MP2-1000N-3755N-1365166074	&	5.792	&	3.97\\
3		&	48x48-100C3-MP2-200C2-MP2-300C2-MP2-400C2-MP2-500N-3755N-1365166209	&	5.625	&	2.15\\
4		&	48x48-100C3-MP2-200C2-MP2-300C2-MP2-400C2-MP2-1000N-3755N-1325085896	&	5.951	&	2.54\\
5		&	48x48-100C3-MP2-200C2-MP2-300C2-MP2-400C2-MP2-1000N-3755N-1325085943	&	6.114	&	2.54\\
6		&	48x48-100C3-MP2-200C2-MP2-300C2-MP2-400C2-MP2-1000N-3755N-1325086048	&	6.339	&	2.54\\
7		&	48x48-100C3-MP2-200C2-MP2-300C2-MP2-400C2-MP2-500N-3755N-1341137514	&	5.995	&	2.14\\
\bottomrule
\end{tabular}
\end{center}
\end{table}
\section{Results}

We built nine MCDNN (Table~\ref{tab:models}) from the eight previously trained nets. Four of them are basic DNN with only one column. We submitted these simple DNN to the competition, too,  because we were interested in their performance---initially we could not access the test set to check them by ourselves, but now we can list them for completeness. Before the deadline, we had to select two models as official competition candidates. Using the validation results, we chose MCDNN 2 and 8. They are also the best on the competition test set.

\begin{table}[h!]
\caption{Recognition errors for our eight MCDNN submitted to ICDAR 2013. DNN architectures are detailed in Table~\ref{tab:net_architectures}. MCDNN structure is described in the middle of the table: X means that the DNN to the left is part of the top MCDNN. {\em First} and {\em Best 10} represent errors computed using the highest and the 10 highest output neuron activation values.
\label{tab:models}} 
\begin{center}
\small
\begin{tabular}{c|ccccccccc}
\toprule
&\multicolumn{8}{c}{MCDNN}\\
		&	0		&	1	&	2	&	3	&	4	&	5	&	6	&	7	&	\textbf{8}\\
\midrule
DNN	0	&	X	&		&	X	&		&		&		&		&	X	&	X\\
DNN	1	&		&	X	&	X	&		&		&		&		&	X	&	X\\
DNN	2	&		&		&	X	&		&		&		&		&		&	X\\
DNN	3	&		&		&	X	&		&		&		&		&		&	X\\
DNN	4	&		&		&		&	X	&		&	X	&	X	&	X	&	X\\
DNN	5	&		&		&		&		&	X	&	X	&	X	&	X	&	X\\
DNN	6	&		&		&		&		&		&		&	X	&		&	X\\
DNN	7	&		&		&		&		&		&		&	X	&		&	X\\
\midrule
&\multicolumn{8}{c}{Speed [ms/character]}\\
			&	3.03&3.03	&12.18	&2.54	&2.54	&5.08	&9.76	&11.14	&22.04\\
\midrule
&\multicolumn{8}{c}{Error [\%]}\\
First	&	5.528	&	5.931	&	4.347	&	5.951	&	6.114	&	5.113	&	4.664	&	4.449	&	\textbf{4.215}\\
Best 10	&	0.396	&	0.425	&	0.291	&	0.461	&	0.457	&	0.387	&	0.340	&	0.303	&	\textbf{0.291}\\
\bottomrule
\end{tabular}
\end{center}
\end{table}

MCDNN always significantly improve over single DNN. The best MCDNN has 4.215\% error, much lower than the best DNN error, 5.528\%. This is an absolute reduction of 1.313\% and a relative reduction of 23.75\%, in line with our observations for other datasets \cite{Ciresan:2012b}.

The competition organizers experimentally measured human error rate as 3.87\%. Our best MCDNN came close: 4.21\% error. Considering the top ten predictions, this MCDNN also has a new record-breaking error rate of 0.291\%, which will be important for more complex context-driven systems using linguistic models.

Despite its size, the best MCDNN can classify 45 characters per second on a single NVIDIA GTX 580. Running on all four cores of an Intel Core i5 2400 3.1GHz, the same MCDNN is 14.29 times slower, requiring 315ms per character. Further speedups can be obtained by optimizing the code for this particular problem or by using more GPUs and/or CPUs.

\section{Conclusions and future work}
Although there are 3755 classes of handwritten Chinese characters, our MCDNN can classify them with almost human performance. They are nearly one fifth better than the best previous artificial method.
Recognition speed on GPUs is high, and scales linearly with their number. A thorough error analysis by native speakers/writers (none of us speaks Chinese) could help to show if there is still room for improvement, or if the remaining errors are just due to illegible characters. Even without additional context-driven linguistic models (which will further reduce errors), our method is ready for practical applications.


\section*{Acknowledgments}This work was partially supported by the \emph{Supervised Deep / Recurrent Nets} SNF grant, Project Code 140399.

{
\bibliographystyle{plain}
\bibliography{bib}
} 
 
\end{document}